%% file: ms.tex
\documentclass[letterpaper, 10 pt, conference]{ieeeconf}  

\IEEEoverridecommandlockouts                              

\usepackage[utf8]{inputenc}
\usepackage[T1]{fontenc}
\usepackage{textcomp}
\usepackage{graphicx} 
\usepackage{todonotes}
\usepackage{pgfplots}
\pgfplotsset{compat=newest}
\usepackage{subfig}
\usepackage{multirow}
\usepackage{tabularx}
\usepackage[hidelinks]{hyperref}
\usepackage{array}
\usepackage{float}
\usepackage{graphbox}
\usepackage{tikz,tikz-3dplot}
\usetikzlibrary{arrows,arrows.meta,automata,backgrounds,calc,chains,%
decorations.markings,decorations.pathreplacing,decorations.pathmorphing,%
matrix,positioning,shapes,shapes.geometric,shapes.symbols,spy,trees,tikzmark}
\usepackage{flushend}
\usepackage[binary-units=true,product-units=single,per-mode=symbol,range-units=single,detect-all]{siunitx}
\DeclareSIUnit\pixel{px}
\usepackage{amsmath} 
\usepackage{amssymb}  

\newcommand{\reffig}[1]{Fig.~\ref{#1}}

\newcommand{\refsec}[1]{Sec.~\ref{#1}}

\newcommand{\etal}{et al.~}
\newcommand{\citep}[1]{(\cite{#1})}

\newcommand{\wrt}{~w.r.t.~}

\usepackage[font=small]{caption}

\makeatletter
\tikzset{
from/.style args={#1 to #2}{
        above right={0cm of #1},
        /utils/exec=\pgfpointdiff
            {\tikz@scan@one@point\pgfutil@firstofone(#1)\relax}
            {\tikz@scan@one@point\pgfutil@firstofone(#2)\relax},
        minimum width/.expanded=\the\pgf@x,
        minimum height/.expanded=\the\pgf@y}}
\makeatother

\title{\LARGE \bf
Autonomous Fire Fighting\\with a UAV-UGV Team at MBZIRC 2020
}

\author{Jan Quenzel$^{*}$, Malte Splietker$^{*}$, Dmytro Pavlichenko, Daniel Schleich\\Christian Lenz, Max Schwarz, Michael Schreiber, Marius Beul, and Sven Behnke%
\thanks{\hspace{-2.2ex}$^{*}$: equal contribution.}%
\thanks{This work has been supported by a grant of the Mohamed Bin Zayed International Robotics Challenge (MBZIRC) and the German Federal Ministry of Education and Research (BMBF) in the project ``Kompetenzzentrum: Aufbau des Deutschen Rettungsrobotik-Zentrums (A-DRZ)``}%
\thanks{Institute for Computer Science VI, Autonomous Intelligent Systems, University of Bonn, Friedrich-Hirzebruch-Allee 8, 53115 Bonn, Germany,
		{\tt\small \{quenzel, \ldots, behnke\}@ais.uni-bonn.de}%
}
}

\begin{document}

\maketitle
\thispagestyle{empty}
\pagestyle{empty}

\begin{abstract}
Every day, burning buildings threaten the lives of occupants and first responders trying to save them. Quick action is of essence, but some areas might not be accessible or too dangerous to enter. Robotic systems have become a promising addition to firefighting, but at this stage, they are mostly manually controlled, which is error-prone and requires specially trained personal.

We present two systems for autonomous firefighting from air and ground we developed for the Mohamed Bin Zayed International Robotics Challenge (MBZIRC) 2020. The systems use LiDAR for reliable localization within narrow, potentially GNSS-restricted environments while maneuvering close to obstacles. Measurements from LiDAR and thermal cameras are fused to track fires, while relative navigation ensures successful extinguishing.

We analyze and discuss our successful participation during the MBZIRC 2020, present further experiments, and provide insights into our lessons learned from the competition.
\end{abstract}


\section{Introduction}
\label{sec:Introduction}
\input{introduction.tex}

\section{Related Work}
\label{sec:Related_Work}
\input{related_work.tex}

\section{UAV Solution}
\label{sec:uav_solution}
\input{uav_solution.tex}

\section{UGV Solution}
\label{sec:ugv_solution}
\input{ugv_solution.tex}

\section{Evaluation}
\label{sec:Evaluation}
\input{evaluation.tex}

\section{Lessons Learned}
\label{sec:lessons_learned}
\input{lessons_learned.tex}

\section{Conclusion}
\label{sec:Conclusion}
\input{conclusion.tex}




\bibliographystyle{IEEEtran}
\bibliography{literature}
\vfill
\end{document}

%% file: introduction.tex
Robotic systems exhibit an immense potential for future applications in search and rescue scenarios~\cite{nex2014review} to aid first responders, overtake life-threatening tasks, or improve accessibility. 
For example, when firefighters had to retreat from within the \textit{Notre-Dame de Paris} cathedral fire due to structural damage, an unmanned ground vehicle (UGV) took over~\cite{NotreDame}. Fire brigades increasingly use remotely controlled unmanned aerial vehicles (UAV) equipped with thermal and color cameras for monitoring tasks~\cite{NIFTI} and for guidance during extinguishing, even without direct line of sight or through smoke. However, latency or communication loss to a remote-controlled robot, e.g., inside a building, can further aggravate the disaster.

\begin{figure}[t]
\centering
\includegraphics[width=\linewidth]{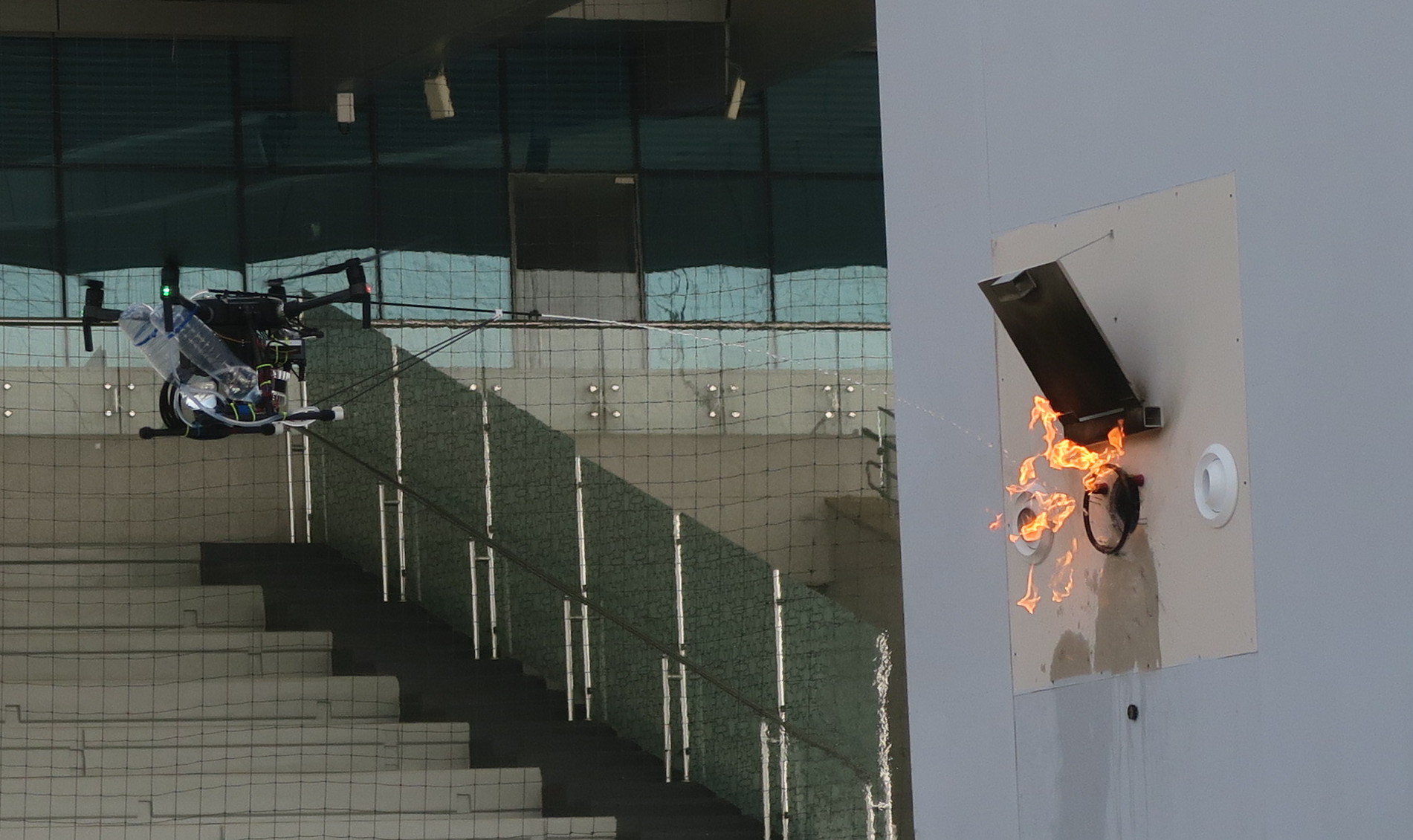}\\
\vspace{0.7em}
\includegraphics[height=3cm]{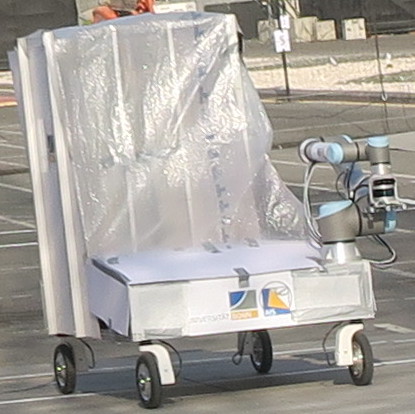}\hfill
\includegraphics[height=3cm]{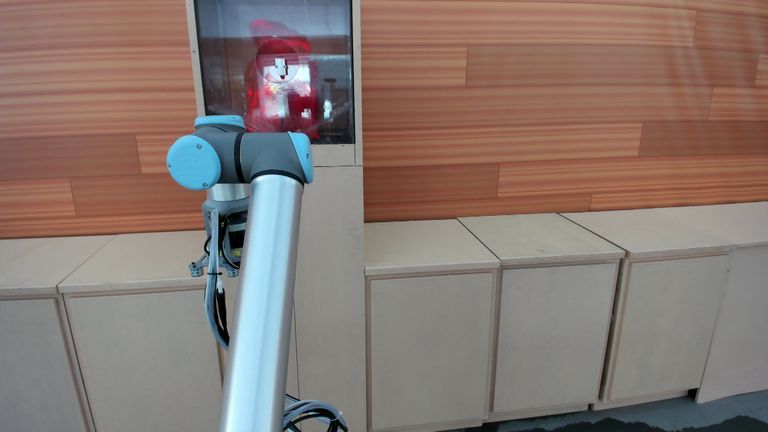}
\caption{Team of fire fighting UAV and UGV at MBZIRC 2020.}
\label{fig:teaser}
\end{figure}

To address these issues, Challenge \num{3} of the Mohamed Bin Zayed International Robotics Challenge (MBZIRC) 2020 \cite{MBZIRC2020} targeted an urban firefighting scenario for autonomous robots to foster and advance the state-of-the-art in perception, navigation, and mobile manipulation. Participants had to detect, approach, and extinguish multiple simulated fires around and inside a building with up to \num{3} UAVs and one UGV. Each fire provided a \SI{15}{\centi\meter} circular opening with a heated plate recessed about \SI{10}{\centi\meter} on the inside. Holes on the outside facade were surrounded by a propane fire ring, while indoor fires had a moving silk flame behind a heating element.

In this work, we present our approach with an autonomous firefighting UAV-UGV team (see \reffig{fig:teaser}) to solve this challenge. In addition to describing and analyzing our integrated systems for solving the tasks, we discuss lessons learned and our technical contributions, including

\begin{itemize}
 \item a precise LiDAR registration method that is applicable both for UAVs and UGVs,
 \item robust 3D fire localization using thermal cameras and LiDAR, and
 \item thermal servoing for navigation relative to a fire.
\end{itemize}

%% file: related_work.tex
Cooperative monitoring and detection of forest and wildfires with autonomous teams of UAVs~\cite{BailonRuiz2020icuas} or UGVs~\cite{Ghamry2016icuas} gained significant attention in recent years~\cite{Delmerico2019jfr}. 
While UGVs can carry larger amounts of extinguishing agents or drag a fire hose~\cite{Liu2016icac}, payload limitations impede the utility of UAVs. 
Hence, Aydin \etal\cite{Aydin2019drones} investigated the deployment of fire-extinguishing balls by a UAV.
Ando \etal\cite{Ando2018ral} developed an aerial hose pipe robot using steerable pressurized water jets for robot motion and fire extinguishing.

In urban environments, thermal mapping~\cite{cho2015survey} is commonly used for building inspection and relies on simultaneously captured color and thermal images from different poses and employs standard photogrammetry pipelines. Real-time assistance for firefighters is provided by Sch\"{o}nauer \etal\cite{schonauer2013live} via thermal augmentation of live images within room-scale environments.

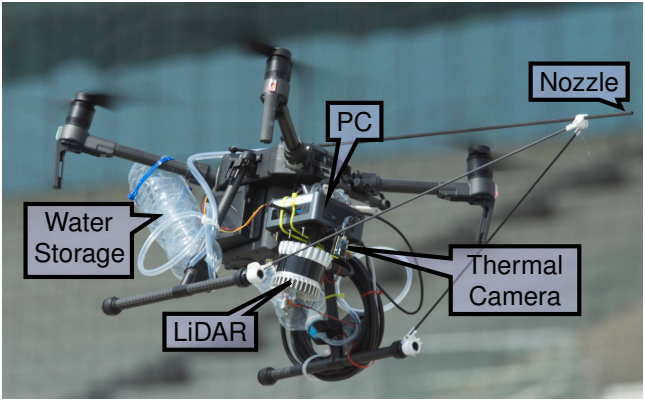
\begin{figure}[t]
  \centering
  \resizebox{1.0\linewidth}{!}{\input{uav.pgf}}
  \caption{Hardware design of our fire-fighting UAV.}
  \label{fig:UAV}
\end{figure}

In contrast, Borrmann \etal\cite{borrmann2013thermal} mounted a terrestrial LiDAR as well as a thermal imager and color camera on a UGV to obtain colorized point clouds, enabling automated detection of windows and walls~\cite{demisse2015interpreting}.
Fritsche \etal\cite{fritsche2017fusion} cluster high-temperature points from fused 2D LiDAR and mechanical pivoting radar to detect and localize heat sources.
Similarly, Rosu \etal\cite{Rosu2019ssrr} acquire a thermal textured mesh using a UAV equipped with LiDAR and thermal camera and estimate the 3D position and extent of heat sources.

New challenges arise where the robots have to operate close to structures. UAVs and UGVs are often equipped with cameras and are remote-controlled by first responders. In contrast, autonomous execution was the goal for Challenge 3 of the MBZIRC 2020.

Team Skyeye~\cite{SkyeyeMBZIRC} used a 6-wheeled UGV and DJI Matrice 600 and 210 v2 UAVs equipped with color and thermal cameras, GPS and LiDAR. A map is prebuilt from LiDAR, IMU, and GPS data to allow online Monte Carlo localization and path planning with Lazy Theta$^{*}$. Fires are detected via thresholding on thermal images. The fire location is estimated with an information filter from either projected LiDAR range measurements or the map.
A water pump for extinguishing is mounted on a pan-tilt-unit on the UGV while being fixed in place on the UAV. 

Although our general approach is similar to Team Skyeye, we rely more heavily upon relative navigation for aiming the extinguisher at the target after initial detection and less on the quality of our map and localization. In comparison, the nozzle's placement on the end effector of the robot arm on our UGV gives us a better reach, and the protruding nozzle on our UAV allows for a safer distance from fire and wall while aiming. Furthermore, where common distance measurements are unavailable, we use the known heat source size for target localization. We detect the holes of outdoor facade fires in LiDAR measurements and fuse these with thermal detections, which allows us to spray precisely on target, where thermal-only aiming would be offset by the surrounding flames.

%% file: uav.pgf
\begin{tikzpicture}
      [boxstyle/.style={font=\sffamily,black,fill=blue!20!white,fill opacity=0.4,text opacity=1,text=black,draw,ultra thick,align=center,rectangle callout}]
      \node[anchor=south east, inner sep = 0] (left_image) at (0,0) {\includegraphics[width=\linewidth]{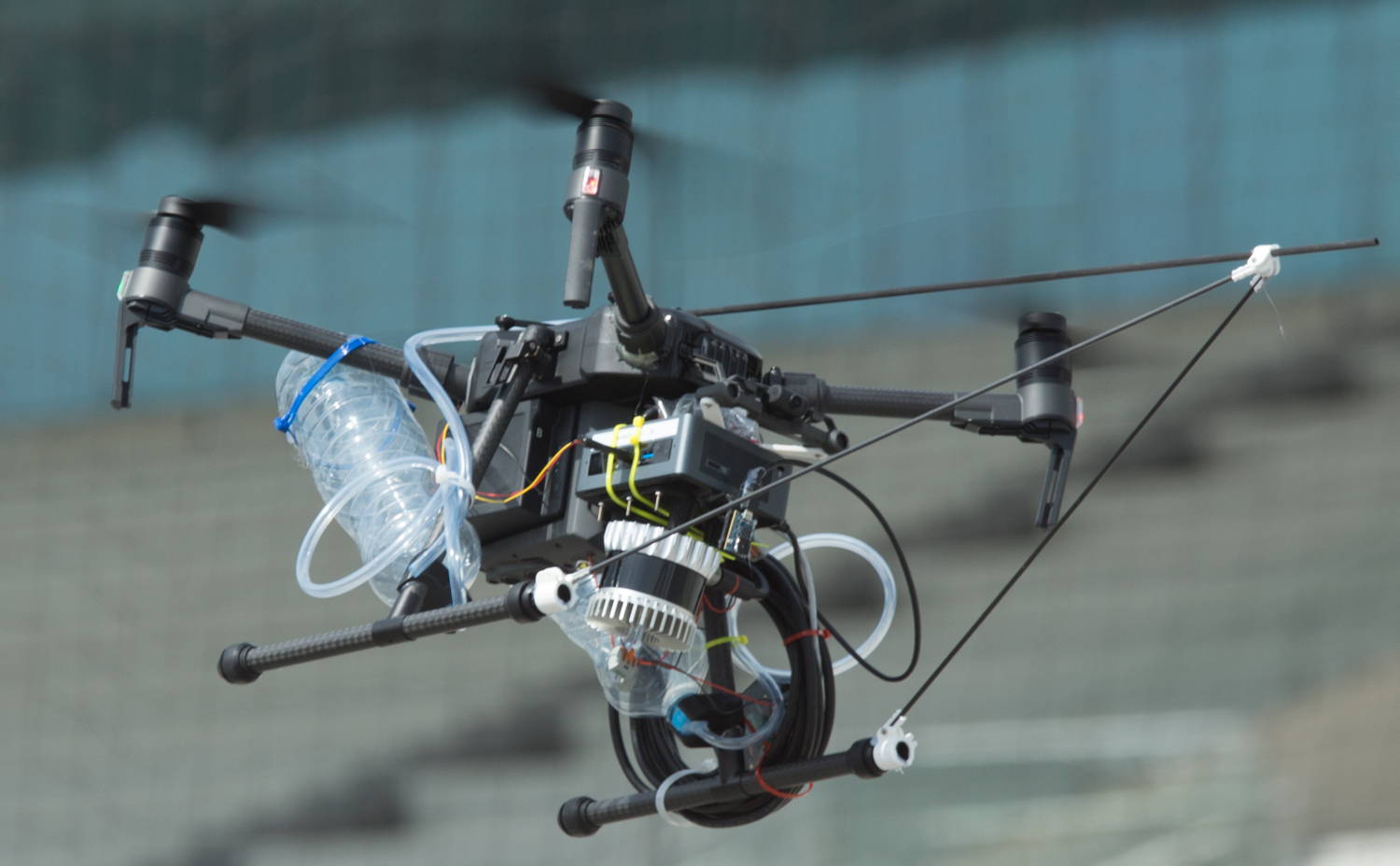}};
    \begin{scope}[shift=(left_image.south west),x={(left_image.south east)},y={(left_image.north west)}]
        \node[boxstyle,callout relative pointer={(-0.03, -0.14)}] at (0.55,0.7) {PC};
        \node[boxstyle,callout relative pointer={(0.08, 0.08)}] at (0.325,0.17) {LiDAR};
        \node[boxstyle,text width=1.5cm,callout relative pointer={(-0.16, 0.05)}] at (0.8,0.3) {Thermal\\Camera};
        \node[boxstyle,text width=1.2cm,callout relative pointer={(0.04, 0.02)}] at (0.12,0.4) {Water\\Storage};
        \node[boxstyle,callout relative pointer={(0.02, -0.02)}] at (0.9,0.8) {Nozzle};
    \end{scope}
\end{tikzpicture}

%% file: uav_solution.tex
We decided to build upon our experience from the first MBZIRC competition~\cite{MBZIRCc3} and equipped commercially available DJI UAVs with sensors and computing power for easy adaptation to task-specific requirements. The limited payload of the previously used DJI M100 did not allow transporting sufficient water for extinguishing the fires. Hence, we upgraded to the more powerful DJI Matrice 210 v2.

\subsection{Hardware Design}
\reffig{fig:UAV} shows the setup of our UAV. It is equipped with an Intel Bean Canyon NUC8i7BEH with Core i7-8559U processor and \SI{32}{\giga\byte} of RAM. For perception and localization of the fires, we combine an Ouster OS1-64 LiDAR and a FLIR Lepton 3.5 thermal camera with \SI{160 x 120}{\pixel} resolution. The M210 provides GNSS-localized ego-motion estimates. We had to deactivate the obstacle avoidance in the forward direction to get close enough to fires for extinguishing.

\begin{figure}[t]
  \centering
  \resizebox{1.0\linewidth}{!}{\input{system.pgf}}
  \caption{UAV system overview.}
  \label{fig:System}
\end{figure}
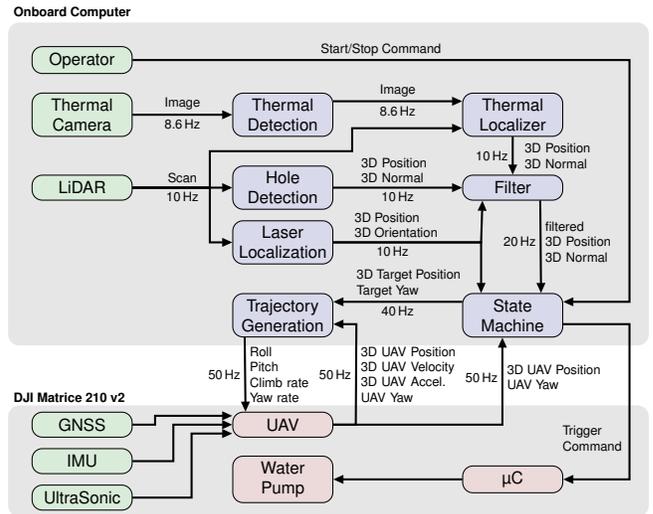

Our water supply is stored in two downward-facing \SI{1.5}{\liter} PET-bottles attached between the rear frame arms and the landing gear. The screw caps are connected via a flexible hose to a windscreen washer pump attached on top of the landing gear. We trigger the pump via an Arduino Nano microcontroller.
We mounted a carbon tube on top of the UAV as an extended nozzle. It is supported by two additional carbon struts from the landing gear. The high location compensates for the height difference in the water jet parabola, allows to perceive the fire with the sensors below, and prevents the water from flowing out on its own during flight maneuvers. We chose a \SI{10}{\degree} downturn for the nozzle, LiDAR, and thermal camera to fly above the fire while maintaining observability. \reffig{fig:System} gives an overview of our UAV software.

\subsection{Perception}

\subsubsection{Laser Localization}
\label{sec:LaserLoc}
GNSS-based localization is subject to local position drift and unreliable close to structures. The usage of RTK-/D-GPS was allowed but penalized and thus not used by us. Hence, we localize our UAV\wrt the building using LiDAR. In a test run, we collected data to generate a Multi-Resolution Surfel Map (MRSMap) of the arena \cite{DavidCTSLAM}.
We predefined the origin of our "field" arena coordinate system in GPS coordinates and transformed the map accordingly. 
During localization, we register the current LiDAR point cloud against the map.
The UAV provides an estimate of its ego-motion from onboard sensors, which we use to update the initial pose. After registration, we update an offset transformation that compensates for the local position drift of the GPS. The translation between consecutive offset updates is bounded to be below \SI{30}{\centi\meter} to prevent potential jumps from incorrect registration.
Although the UAV ego-motion estimate is reliable in the short term, a continuous update is necessary. We experienced incorrect ego-motion estimates when approaching the building at specific heights. This issue is likely due to the ultra-sonic sensor measuring the distance to the building rather than the height above the ground.
All subsequent navigation poses are expressed in the localized map frame.

\subsubsection{Hole Detection}
\begin{figure}[t]
  \centering
  \resizebox{\linewidth} {!}
  {
    \begin{tikzpicture}
      \node[anchor=south east] at (0,0) {\includegraphics[width=\linewidth]{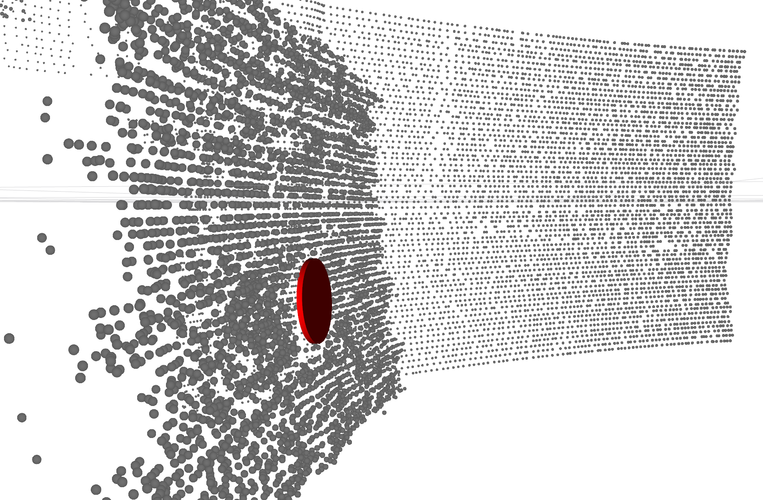}};
        \node[anchor=south east] at (0,0) {\includegraphics[width=.4\linewidth]{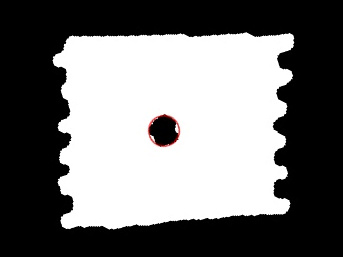}};
    \end{tikzpicture}
  }
  \caption{Detected hole (red disk) inside a point cloud. The lower-right corner shows the points of the left plane projected into the virtual camera after morphological transformation as well as the detected circle.}
  \label{fig:hole_detection}
\end{figure}

The water jet exiting the nozzle is very narrow; thus, precise aiming is required in order to extinguish a fire with a limited amount of water. As the global tracking accuracy is not sufficient for this task, we apply a hole detection algorithm on the LiDAR point clouds for relative navigation.
First, we extract planes using RANSAC. For every such plane, we project the contained points into a virtual camera perpendicular to the plane. We apply morphological transformations to the resulting image in order to close gaps between projected points. Shape detection is used on the image to find potential holes. After re-projecting the circles into 3D, overly large or small holes are discarded, as the scenario specifies a constant \SI{150}{\milli\meter} diameter for all target holes. An example is shown in~\reffig{fig:hole_detection}.

\subsubsection{Thermal Detection and Heat Source Localization}
\label{ssec:uav_thermal_detection}
Our thermal detector applies lower and upper bounds to threshold the intensity. We find contours in the thresholded image. For each contour, we compute the area, bounding box, center of intensity, as well as min, max, and mean intensity for further processing. Only contours of a specific size are retained.

The hole detection provides a reliable estimate at close range (\textless\SI{3}{\meter}) but fails to detect the opening from multiple meters away where the heat source is already visible in the thermal image and we have distance measurements on the wall. Hence, to obtain an approximate heat source position, we project the LiDAR point cloud into the thermal image. We use all points falling into the bounding box of the detected heat contour and calculate their mean position and normal from the covariance. Since hole and thermal detections on an equivalent heating element were very stable, we used them to calibrate the transformation between LiDAR and thermal camera. We recorded a heat source from multiple positions and minimized the reprojection error between the center of intensity of the heat source and the projected mean position of the detected hole.
The resulting accuracy is sufficient to successfully approach the target.

\subsubsection{Filter}
Our thermal and hole detector both output lists of detections $d_i=(p_i,n_i)$, consisting of a position $p_i$ and a normal $n_i$.
A filtering module processes these detections to reject outliers and combines both detection types into an estimate of the position and orientation of the currently tracked fire.
To do this, we keep track of a history $\mathcal H = ( (p_1, n_1), \dots, (p_{10}, n_{10}) )$ of ten recent valid detections and estimate the target position $p^\star = \frac{1}{|\mathcal H|}\sum_{i = 1}^{|\mathcal H|} p_i $ and normal $n^\star := \frac{1}{|\mathcal H|}\sum_{i = 1}^{|\mathcal H|} n_i $ as running averages over the detection history.
Mind that the detection history may contain both thermal and hole detections.
Thermal detections are necessary to determine which of the possible targets is currently on fire. However, we found that hole detections give a more precise estimation of the target position and especially of its orientation. Hence, we use thermal detections to initialize the target estimate and subsequently update it using hole detections if available.
In the following, we detail the initialization process and the estimate update during target tracking:

\paragraph{Target Initialization}
For the initialization process, we determine the feasibility of thermal detections by checking the angle between the detected normal and the line-of-sight from the UAV position to the detected fire position.
If this angle exceeds \SI{45}{\degree}, the detection is rejected.
The remaining detections are collected until ten out of the last twenty detected heat positions lie within a distance of \SI{1}{\meter}.
These ten detections are used to initialize the detection history $\mathcal H$ more robustly against outliers.

\paragraph{Target Tracking}
We filter thermal, and hole detections based on the same feasibility checks as above: detections for which the angle between the detected normal and the line-of-sight exceeds \SI{45}{\degree} are discarded.
To remove outliers, we additionally only consider detections which positions lie within a ball of radius \SI{1}{\meter} around the current position estimate and for which the angle between the detected normal and the estimated normal is lower than \SI{45}{\degree}.

Although we keep track of the most recent thermal detection, we only add it to the detection history $\mathcal H$ if there has not been a valid hole detection within the last second.
Thus, we ensure that we are still able to estimate target positions if the hole detector fails, but otherwise only use the more precise information from hole detections.
In the case of multiple holes close to the target position, the estimate might drift away from the target if we only use hole detections.
Furthermore, we have to recover from situations in which the initial heat detections were wrong or in which the fire has been extinguished in the meantime.
To address these issues, we only add hole detections to the detection history $\mathcal H$ if there has been a heat detection within the last second and the detected hole position lies within a radius of \SI{1}{\meter} around the latest heat detection.
If no detection was added to the history for more than \SI{2}{\second}, tracking times out, and we re-initialize the target estimate.

\subsection{Control}
\begin{figure}[t]
	\centering
	\resizebox{1.0\linewidth}{!}{\input{uav_state_machine.pgf}}
	\caption{Flowchart of the UAV state machine.}
	\label{fig:uav_state_machine}
\end{figure}
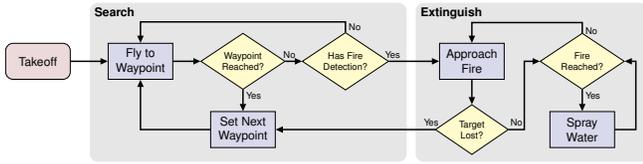

We build upon our experience from the MBZIRC 2017 and reuse the method of Beul and Behnke~\cite{beul2017icuas} for trajectory generation and low-level control.
The high-level control of the UAV is performed by a Finite State Machine (FSM)\footnote{Open-source: \url{https://github.com/AIS-Bonn/nimbro_fsm2}}. The FSM uses inputs from components described above to produce navigation waypoints to locate and approach fires, as well as to control the water spraying during extinguishing. It also ensures that the UAV stays within arena limits and altitude corridor. The diagram of the FSM is shown in~\reffig{fig:uav_state_machine}.

\textbf{\textit{Search} state:}
The UAV flies around the building in order to detect and localize a fire. The route around the building is defined as a linked list of waypoints. Each waypoint is defined in the building frame, obtained from laser localization. The route forms a convex hull around the building's XY plane. Thus, transfer between two consequent waypoints can be safely performed without additional planning for collision avoidance, simplifying the overall system. While moving between waypoints, the detection and filter pipeline collects data. There are two types of waypoints: {\it Transfer} and {\it Observation}. Upon arrival at a {\it Transfer} waypoint, the UAV proceeds to the next waypoint immediately. When arriving at an {\it Observation} waypoint located in front of the known hole locations, the UAV hovers for a predefined duration of \SI{2}{\second}, looking for fires. The UAV switches into the \textit{Extinguish} state if at any time a detection satisfies the following conditions:

\begin{enumerate}
	\item The detection was observed at least five times in a row,
	\item the detection Z coordinate lies within the allowed altitude interval, and
	\item the angle between UAV heading and detection heading does not exceed the predefined $\SI{45}{\degree}$ limit.
\end{enumerate}

\textbf{\textit{Extinguish} state:}
The purpose of this state is to arrive at the extinguishing position without losing the detected fire on the way. Given the target from the previous state, the navigation goal is set at a predefined offset from the fire in the direction of its estimated normal (\SI{2.1}{\meter} away and \SI{0.35}{\meter} above). This offset was determined empirically to suit the parabola of our water spraying system. The goal heading of the UAV at each time step is modified by adding an angle between the current heading of the UAV and the heading of the fire. By doing so, we try to continuously keep the fire in the center of the thermal camera's field of view during the approach. This state performs relative navigation, with the UAV moving relative to the currently observed fire without relying on laser localization. This allows to achieve a precise placement of the UAV relative to the fire, which is necessary for consequent extinguishing. Once the UAV reaches the goal pose, the pump is started. It is stopped as soon as UAV deviates from the current goal pose more than a predefined threshold for either position or heading. The UAV keeps track of the fire and updates the goal pose accordingly. If the UAV does not receive new detections of the fire for longer than \SI{5}{\second}, the extinguishing is aborted, and it switches back to the {\it Search} state. If the UAV did not lose the target while extinguishing, the pump is stopped after all water has been sprayed. In that case, the UAV returns and hovers above the start position, waiting for manual landing and refueling of the water storage.

For additional safety, at all times, the building frame from laser localization is continuously tracked. If a sudden jump of more than \SI{1}{\meter} is detected, the UAV immediately transitions into the {\it Stop} state, aborting any ongoing operation, and hovers at its current position until a safety pilot takes over control.

%% file: system.pgf
\begin{tikzpicture}
[content_node/.append style={font=\sffamily,minimum size=1.5em,minimum width=6em,draw,align=center,rounded corners,scale=0.65},
label_node/.append style={font=\sffamily,scale=0.5},
group_node/.append style={font=\sffamily,dotted,align=center,rounded corners,inner sep=1em,thick},>={Stealth[inset=0pt,length=4pt,angle'=45]}]

\definecolor{red}{rgb}     {0.5,0.0,0.0}
\definecolor{green}{rgb}   {0.0,0.5,0.0}
\definecolor{blue}{rgb}    {0.0,0.0,0.5}
\definecolor{grey}{rgb}    {0.5,0.5,0.5}

\draw[thick, rounded corners, grey!20!white,fill] (-4.0,0.6) -- (4.75,0.6) -- (4.75,5.0) -- (-4.0,5.0) -- cycle;
\draw[thick, rounded corners, grey!20!white,fill] (-4.0,-1.75) -- (4.75,-1.75) -- (4.75,-0.25) -- (-4.0,-0.25) -- cycle;

\node(Operator)[content_node,fill=green!15!white] at (-3.0,4.5) {Operator};
\node(Thermal_Camera)[content_node,fill=green!15!white] at (-3.0,3.75) {Thermal\\Camera};
\node(LIDAR)[content_node,fill=green!15!white] at (-3.0,2.75) {LiDAR};
\node(Thermal_Detection)[content_node,fill=blue!15!white] at (-.25,3.75) {Thermal\\Detection};
\node(Hole_Detection)[content_node,fill=blue!15!white] at (-.25,2.75) {Hole\\Detection};\node(Laser_Localization)[content_node,fill=blue!15!white] at (-.25,2) {Laser\\Localization};
\node(Thermal_Localizer)[content_node,fill=blue!15!white] at (2.9,3.75) {Thermal\\Localizer};

\node(Filter)[content_node,fill=blue!15!white] at (2.9,2.75) {Filter};
\node(State_Machine)[content_node,fill=blue!15!white] at (2.9,1) {State\\Machine};
\node(Trajectory_Generation)[content_node,fill=blue!15!white] at (-.25,1) {Trajectory\\Generation};

\node(UAV)[content_node,fill=red!15!white] at (-.25,-0.5) {UAV};
\node(GNSS)[content_node,fill=green!15!white] at (-3.0,-0.5) {GNSS};
\node(IMU)[content_node,fill=green!15!white] at (-3.0,-1.0) {IMU};
\node(UltraSonic)[content_node,fill=green!15!white] at (-3.0,-1.5) {UltraSonic};
\node(UC)[content_node,fill=red!15!white] at (2.9, -1.25) {µC};
\node(Pump)[content_node,fill=red!15!white] at (-.25,-1.25) {Water\\Pump};

\draw[->, thick] (Thermal_Camera) -- node[label_node,midway,below] {\SI{8.6}{\hertz}} node[label_node,midway,above] {Image} (Thermal_Detection);

\draw[->, thick] (Thermal_Detection.15) -- node[label_node,midway,below] {\SI{8.6}{\hertz}} node[label_node,midway,above] {Image} (Thermal_Localizer.165);

\draw[->,thick] (LIDAR) -- (LIDAR -| -1.25,-0.75) -- (-1.25,3.25) -- (0.7,3.25 )  -- ( 0.7,0 |-Thermal_Localizer.195) -- (Thermal_Localizer.195);
\draw[->,thick] (LIDAR) -- (LIDAR -| -1.25,-0.75) -- (-1.25,1.0 |- Laser_Localization.180)
  -- (Laser_Localization);
\draw[->, thick] (LIDAR) -- node[label_node,midway,below] {\SI{10}{\hertz}} node[label_node,midway,above] {Scan} (Hole_Detection);

\draw[->, thick] (Thermal_Localizer) -- node[label_node,midway,left] {\SI{10}{\hertz}}
 node[label_node,midway,right,text width=2cm,xshift=0.2cm] {3D~Position 3D~Normal} (Filter);
\draw[->, thick] (Hole_Detection) -- node[label_node,midway,below] {\SI{10}{\hertz}}
 node[label_node,midway,above,text width=2cm] {3D~Position 3D~Normal}(Filter);

\draw[->, thick] (Laser_Localization) -- (Laser_Localization -| State_Machine.145) --  (Filter.203);
\draw[->, thick] (Laser_Localization) -- node[label_node,midway,below,xshift=-0.4cm] {\SI{10}{\hertz}} node[label_node,midway,above,align=left,xshift=-0.3cm] {3D~Position\\3D~Orientation} ( Laser_Localization -| State_Machine.145) -| (State_Machine.145);

\draw[->, thick] (Filter.335) -- node[label_node,midway,right,align=left,text width=2.1cm,yshift=0.15cm] {filtered 3D~Position 3D~Normal} node[label_node,midway,left,align=center,yshift=0.15cm] {\SI{20}{\hertz}} (State_Machine.38);

\draw[->, thick] (State_Machine.165) -- node[label_node,midway,below] {\SI{40}{\hertz}} node[label_node,midway,above,align=left,xshift=0.3cm] {3D~Target~Position\\Target~Yaw} (Trajectory_Generation.15);
\draw[->, thick] (Trajectory_Generation.210) -- node[label_node,midway,left] {\SI{50}{\hertz}} node[label_node,midway,right,text width=1cm] {Roll Pitch Climb~rate Yaw~rate} (UAV.161);

\draw[->, thick] (UAV) -- (0.75,-0.5) -- node[label_node,midway,left,yshift=0.0cm] {\SI{50}{\hertz}} node[label_node,midway,right,text width=2.8cm,yshift=0.0cm] {3D~UAV~Position 3D~UAV~Velocity 3D~UAV~Accel. UAV~Yaw} (0.758,1.0 |- Trajectory_Generation.350 ) -- (Trajectory_Generation.350);
\draw[->, thick] (UAV) -- (State_Machine.245 |- UAV) -- node[label_node,midway,left,yshift=0.1cm] {\SI{50}{\hertz}} node[label_node,midway,right,align=left,yshift=0.1cm] {3D~UAV~Position\\UAV~Yaw}(State_Machine.245);

\draw[->, thick] (GNSS) -- (GNSS -| -2,-1.0)  -- (-2,-1.0 |- UAV.170) -- node[label_node,midway,left] {} node[label_node,midway,right] {} (UAV.170);
\draw[->, thick] (IMU) --  (IMU -| -1.75,-1.0) --  (-1.75,-1.0 |- UAV.180)  -- node[label_node,midway,left] {} node[label_node,midway,right] {} (UAV.180);
\draw[->, thick] (UltraSonic) -- (UltraSonic -| -1.5,-1.0)  -- (-1.5,-1.0 |- UAV.190) -- node[label_node,midway,left] {} node[label_node,midway,right] {} (UAV.190);

\draw[->, thick] (Operator) -- node[label_node,midway,above] {Start/Stop Command} node[label_node,midway,left] {} node[label_node,midway,right] {} (Operator -| 4.5,0 ) -- ( 4.5, 0 |- State_Machine.15) -- ( State_Machine.15 );

\draw[->, thick] (State_Machine.350) -- (State_Machine.350 -| 4.5,0 ) -- node[label_node,midway,left,text width=1.7cm, yshift=-1cm] {Trigger Command} ( 4.5, 0 |- UC) -- (UC);

\draw[->, thick] (UC.180) -- (Pump.0);

\node(ROS_Group_Label)[label_node,anchor=south west] at (-4.0,5.0) {\textbf{Onboard Computer}};
\node(UAV_Group_Label)[label_node,anchor=south west] at (-4.0,-0.25) {\textbf{DJI Matrice 210 v2}};
   
\end{tikzpicture}

%% file: uav_state_machine.pgf
\begin{tikzpicture}[font=\sffamily,on grid,>={Stealth[inset=0pt,length=4pt,angle'=45]}]
\tikzset{every node/.append style={node distance=3.0cm}}
\tikzset{terminal_node/.append style={minimum size=1.5em,minimum height=3em,minimum width={width("Search Point")+0.2em},draw,align=center,rounded corners,scale=0.65}}
\tikzset{content_node/.append style={minimum size=1.5em,minimum height=3em,minimum width={width("Search Point")+0.2em},draw,align=center,scale=0.65,fill=blue!15!white}}
\tikzset{label_node/.append style={scale=0.5, near start}}
\tikzset{group_node/.append style={align=center,rounded corners,inner sep=1em,thick}}
\tikzset{decision_node/.append style={align=center,scale=0.5,shape aspect=1.5,minimum width=7.9em,minimum height=5.4em,diamond,draw,fill=yellow!25!white,font=\sffamily\normalsize,node distance=3.9cm}}

\definecolor{red}{rgb}     {0.5,0.0,0.0}
\definecolor{green}{rgb}   {0.0,0.5,0.0}
\definecolor{blue}{rgb}    {0.0,0.0,0.5}
\definecolor{grey}{rgb}    {0.5,0.5,0.5}

\draw[thick, rounded corners, grey!20!white,fill] (1.0, -1.9) -- (7.0, -1.9) -- (7.0, 1.1) -- (1.0, 1.1) -- cycle;

\draw[thick, rounded corners, grey!20!white,fill] (7.2, -1.9) -- (11.6, -1.9) -- (11.6, 1.1) -- (7.2, 1.1) -- cycle;

\node(takeoff)[terminal_node,fill=red!15!white] at (0, 0) {Takeoff};
\node(fly_to_waypoint)[content_node, right of=takeoff] {Fly to\\Waypoint};
\node(waypoint_reached)[decision_node, right of=fly_to_waypoint] {Waypoint\\Reached?};
\node(set_next_waypoint)[content_node, below of=waypoint_reached, node distance=2.0cm] {Set Next\\Waypoint};
\node(has_fire_detection)[decision_node, right of=waypoint_reached] {Has Fire\\Detection?};
\node(approach)[content_node, right of=has_fire_detection, node distance=3.7cm] {Approach\\Fire};
\node(target_lost)[decision_node, below of=approach, node distance=2.6cm] {Target\\Lost?};
\node(fire_approached)[decision_node, right of=approach, node distance=4.2cm] {Fire\\Reached?};
\node(spray_water)[content_node, below of=fire_approached, node distance=2.0cm] {Spray\\Water};

\draw[->, thick] (takeoff) -- (fly_to_waypoint);
\draw[->, thick] (fly_to_waypoint) -- (waypoint_reached);
\draw[->, thick] (waypoint_reached) -- node[label_node,right] {Yes} (set_next_waypoint);
\draw[->, thick] (waypoint_reached) -- node[label_node,above] {No} (has_fire_detection);
\draw[->, thick] (set_next_waypoint) -| (fly_to_waypoint);
\draw[->, thick] (has_fire_detection.north) -- node[label_node,right, midway] {No} ++(0, 0.2) -| (fly_to_waypoint);
\draw[->, thick] (has_fire_detection.east) -- node[label_node,above] {Yes} ++(0.3, 0) -- (approach);
\draw[->, thick] (approach) -- (target_lost);
\draw[->, thick] (target_lost.west) -- node[label_node,above] {Yes} ++(-0.3, 0) -- (set_next_waypoint);
\draw[->, thick] (target_lost.east) -| node[label_node,above] {No} ++(0.3, 0.0) |- (fire_approached.west);
\draw[->, thick] (fire_approached.north) -- node[label_node,right] {No} ++(0, 0.2) -| (approach);
\draw[->, thick] (fire_approached) -- node[label_node,right] {Yes} (spray_water);
\draw[->, thick] (spray_water.east) -| ++(0.4, 0.0) |- (fire_approached.east);

\node[scale=0.65, anchor=north west] at (1, 1.1) {\textbf{Search}};
\node[scale=0.65, anchor=north west] at (7.2, 1.1) {\textbf{Extinguish}};

\end{tikzpicture}

%% file: ugv_solution.tex
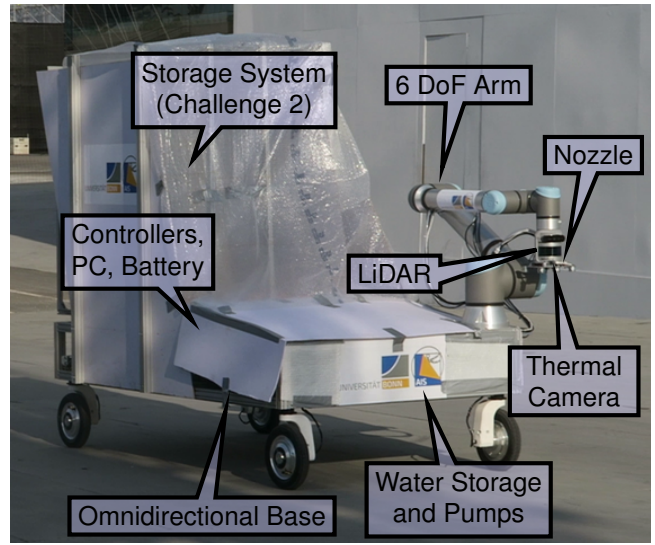
\begin{figure}[t]
  \centering
  \resizebox{1.0\linewidth}{!}{\input{ugv.pgf}}
  \caption{Hardware design of our fire fighting UGV.}
  \label{fig:UGV}
\end{figure}

We mainly developed our ground robot Bob to solve Challenge 2 (pick and place of brick-shaped objects)~\cite{ssrr_ch2} based on our very successful UGV Mario, which won the MBZIRC 2017 competition~\cite{schwarz2019team}.
We added a thermal camera, water storage, and pumps to solve the firefighting challenge.

\subsection{Hardware Design}
The UGV hardware includes a four-wheeled omnidirectional base, a 6-DoF Universal Robots UR10e arm with a Velodyne VLP-16 3D LiDAR and a FLIR Lepton 3.5 thermal camera mounted on the wrist. A water storage for up to \SI{10}{\liter}, and two windscreen washer pumps, are installed in the robot base (see \reffig{fig:UGV}). A standard ATX mainboard with a quad-core Intel Core i7-6700 CPU and \SI{64}{\giga\byte} RAM enables onboard computation. The robot is powered by an eight-cell LiPo battery with \SI{20}{\ampere{}\hour} and \SI{29.6}{\volt} nominal voltage for roughly one hour operation time.
The robot's base has a footprint of \SI{1.9 x 1.4}{\meter} and is driven by four direct-drive brushless DC hub motors. Each wheel is attached via a Dynamixel H54-200-S500-R yaw servo to enable omnidirectional driving capabilities. The two pumps are connected in series. This setup generates enough power to spray roughly \SI{1}{\liter} of water in \SI{10}{\second} over a height difference of up to \SI{1.5}{\meter}.

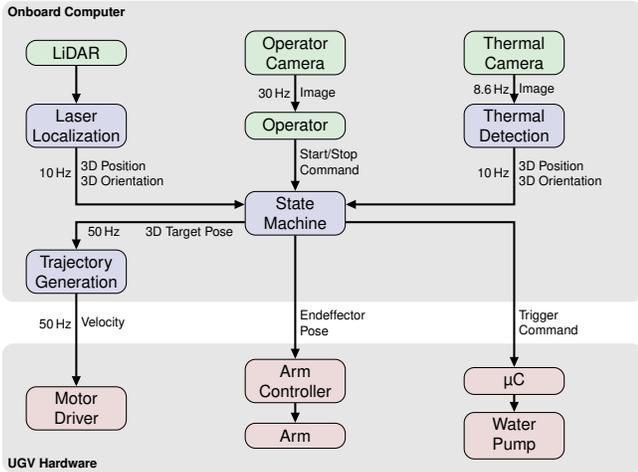
\begin{figure}[t]
  \centering
  \resizebox{1.0\linewidth}{!}{\input{ugv_system.pgf}}
  \caption{UGV system overview.}
  \label{fig:ugv_system}
\end{figure}

\subsection{Perception}
\label{sec:ugv_perception}
For heat source detection, we use the same algorithms as for our UAV. The simulated fires on the ground level are placed inside an acrylic glass enclosure, which brings new challenges:
Firstly, only a fraction of the infrared rays penetrate the walls, so the thermal detection works only through the hole in the front, which results in a very narrow window for detection. Secondly, the LiDAR hardly ever perceives transparent acrylic glass, so hole detection cannot be reused for determining the relative position of the heat source. Assuming the heating element has reached target temperature and is observed at a nearly orthogonal angle through the hole, the bounding box of the thermal detections can be used to estimate the distance. We used an equivalent heating element for calibrating the measurements. Back projecting the detection's center point with this distance results in a 3D position estimate of the heat source.

\begin{figure}[t]
  \centering
  \resizebox{1.0\linewidth}{!}{\input{state_machine.pgf}}
  \caption{The flowchart of our UGV state machine.}
  \label{fig:ugv_state_machine}
\end{figure}
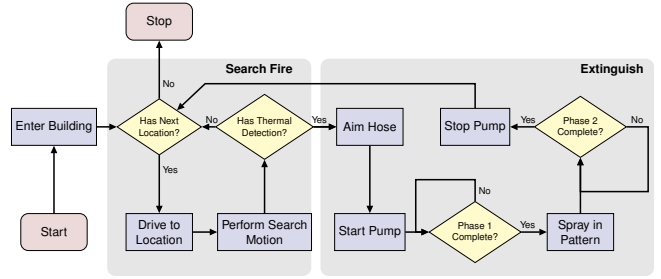

We reused a modified version of the laser-based localization from \refsec{sec:LaserLoc} for our UGV. Instead of estimating a correction transformation from the GPS-based "field" frame to the "map" frame, we calculate the offset transformation between the "field" origin and the "odometry" frame without GPS. The wheel encoder odometry is reset each time at startup but is accurate enough to be used as initialization during incremental scan matching.

The map is combined from two separately recorded segments a) from the start area to the kitchen entrance, and b) from a round trip inside the kitchen. Both segments are mapped as before and then aligned. While approaching the building, we use the outdoor map and seamlessly switch to the indoor map before passing the doorway.

\subsection{High-level Control}

\reffig{fig:ugv_system} gives a system overview. Details for the central state machine are shown in~\reffig{fig:ugv_state_machine}. The UGV navigates from the start zone to the inside of the building using a set of static waypoints. The scenario specifies exactly two static fire locations within the kitchen, one of which is burning, which we utilized by driving the UGV to pre-determined positions roughly in front of the fire and performing a rectangular search motion with the arm. When a heat source is detected, the robot starts the extinguishing procedure. If not, it proceeds to the next location.

The nozzle is aimed so that the water jet will directly hit the heating element (the trajectory was determined experimentally). Since the relative position estimate of the heat source described in \refsec{sec:ugv_perception} might differ from our calibration and vary with distance, the aiming process is performed in multiple \SI{10}{\centi\meter} arm motions to incorporate new measurements. To further increase the chances of successfully extinguishing, we split the actual water spraying into two phases. For the first half of the assigned water, the nozzle is kept still. In the second phase, the remaining water is sprayed in an hourglass-shaped pattern by tilting the nozzle. During the Grand Challenge run, the first phase already delivered a sufficient amount of water to extinguish the fire.
Even though we never observed the thermal detector producing any false-positives during our runs, the robot will continue proceeding to the second fire after extinguishing the first one. The water capacity is more than sufficient, with approximately \SI{4}{\liter} per fire.

%% file: ugv.pgf
\begin{tikzpicture}[boxstyle/.style={font=\sffamily,black,fill=blue!20!white,fill opacity=0.4,text opacity=1,text=black,draw,ultra thick,align=center,rectangle callout}]
  \node[anchor=south east, inner sep = 0] (left_image) at (0,0) {\includegraphics[width=\linewidth]{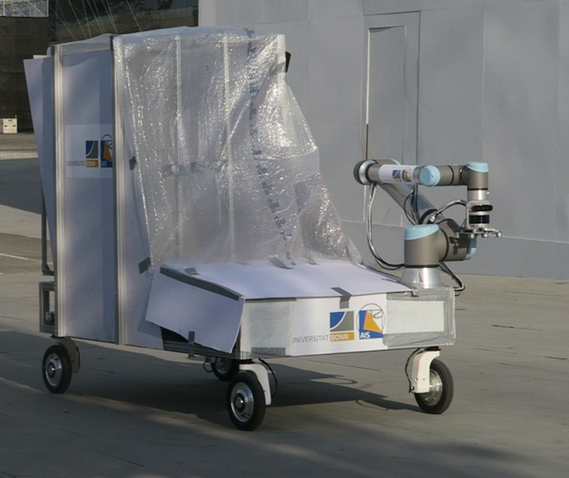}};
  \begin{scope}[shift=(left_image.south west),x={(left_image.south east)},y={(left_image.north west)}]
    \node[boxstyle,callout relative pointer={(0.04, -0.06)}] at (0.2,0.54) {Controllers,\\PC, Battery};
    \node[boxstyle,callout relative pointer={(0.15, 0.03)}] at (0.6,0.5) {LiDAR};
    \node[boxstyle,text width=1.5cm,callout relative pointer={(-0.019, 0.128)}] at (0.88,0.3) {Thermal\\Camera};
    \node[boxstyle,callout relative pointer={(-0.03, 0.11)}] at (0.7,0.08) {Water Storage\\and Pumps};
    \node[boxstyle,callout relative pointer={(-0.032, -0.135)}] at (0.91,0.72) {Nozzle};
    \node[boxstyle,callout relative pointer={(-0.02, -0.12)}] at (0.7,0.85) {6 DoF Arm};
    \node[boxstyle,callout relative pointer={(0.03, 0.16)}] at (0.3,0.05) {Omnidirectional Base};
    \node[boxstyle,callout relative pointer={(-0.03, -0.06)}] at (0.35,0.84) {Storage System\\(Challenge 2)};
  \end{scope}
\end{tikzpicture}

%% file: ugv_system.pgf
\begin{tikzpicture}
[content_node/.append style={font=\sffamily,minimum size=1.5em,minimum width=6em,draw,align=center,rounded corners,scale=0.65},
label_node/.append style={font=\sffamily,scale=0.5},
group_node/.append style={font=\sffamily,dotted,align=center,rounded corners,inner sep=1em,thick},>={Stealth[inset=0pt,length=4pt,angle'=45]}]

\definecolor{red}{rgb}     {0.5,0.0,0.0}
\definecolor{green}{rgb}   {0.0,0.5,0.0}
\definecolor{blue}{rgb}    {0.0,0.0,0.5}
\definecolor{grey}{rgb}    {0.5,0.5,0.5}

\draw[thick, rounded corners, grey!20!white,fill] (-4.0,0.6) -- (4.75,0.6) -- (4.75,4.7) -- (-4.0,4.7) -- cycle;
\draw[thick, rounded corners, grey!20!white,fill] (-4.0,-1.75) -- (4.75,-1.75) -- (4.75,0.0) -- (-4.0,0.0) -- cycle;

\node(Operator)[content_node,fill=green!15!white] at (0.0,3.0) {Operator};
\node(Operator_Camera)[content_node,fill=green!15!white] at (0.0,4.0) {Operator\\Camera};
\node(Thermal_Camera)[content_node,fill=green!15!white] at (3.0,4.0) {Thermal\\Camera};
\node(LIDAR)[content_node,fill=green!15!white] at (-3.0,4.0) {LiDAR};
\node(Thermal_Detection)[content_node,fill=blue!15!white] at (3.0,3.0) {Thermal\\Detection};
\node(Laser_Localization)[content_node,fill=blue!15!white] at (-3.0,3.0) {Laser\\Localization};

\node(State_Machine)[content_node,fill=blue!15!white] at (0.0,1.8) {State\\Machine};
\node(Trajectory_Generation)[content_node,fill=blue!15!white] at (-3.0,1) {Trajectory\\Generation};

\node(Motor_Driver)[content_node,fill=red!15!white] at (-3.0,-0.875) {Motor\\Driver};
\node(Arm_Controller)[content_node,fill=red!15!white] at (0.0,-0.5) {Arm\\Controller};
\node(Arm)[content_node,fill=red!15!white] at (0.0,-1.25) {Arm};
\node(UC)[content_node,fill=red!15!white] at (3.0, -0.5) {µC};
\node(Pump)[content_node,fill=red!15!white] at (3.0,-1.25) {Water\\Pump};

\draw[->, thick] (Thermal_Camera) -- node[label_node,midway,left] {\SI{8.6}{\hertz}} node[label_node,midway,right] {Image} (Thermal_Detection);
\draw[->, thick] (Thermal_Detection) -- node[label_node,midway,left] {\SI{10}{\hertz}} node[label_node,midway,right,align=left] {3D~Position\\3D~Orientation} ++(0, -1) |- (State_Machine.10);

\draw[->,thick] (LIDAR) -- (Laser_Localization);
\draw[->, thick] (Laser_Localization) -- node[label_node,midway,left] {\SI{10}{\hertz}} node[label_node,midway,right,align=left] {3D~Position\\3D~Orientation} ++(0, -1) |- (State_Machine.170);

\draw[->, thick] (State_Machine.190) -- node[label_node,midway,below] {\SI{50}{\hertz}\qquad3D~Target~Pose} (State_Machine.190 -| Trajectory_Generation.90) -- (Trajectory_Generation.90);
\draw[->, thick] (Trajectory_Generation) -- node[label_node,midway,left,yshift=0.45cm] {\SI{50}{\hertz}} node[label_node,midway,right,text width=2cm,yshift=0.45cm] {Velocity} (Motor_Driver);

\draw[->, thick] (Operator_Camera) -- node[label_node,midway,left] {\SI{30}{\hertz}} node[label_node,midway,right] {Image} (Operator);
\draw[->, thick] (Operator) -- node[label_node,midway,right, text width=2.0cm, yshift=0.1cm] {Start/Stop Command} (State_Machine);

\draw[->, thick] (State_Machine.south) -- node[label_node,midway,right,text width=2.0cm, yshift=-0.70cm] {Endeffector Pose} (Arm_Controller); \draw[->, thick] (Arm_Controller) -- (Arm);

\draw[->, thick] (State_Machine.350) -| node[label_node,midway,right,text width=1.7cm, yshift=-2.75cm] {Trigger Command} (UC);
\draw[->, thick] (UC) -- (Pump);

\node(ROS_Group_Label)[label_node,anchor=north west] at (-4.0,4.7) {\textbf{Onboard Computer}};
\node(UGV_Group_Label)[label_node,anchor=south west] at (-4.0,-1.75) {\textbf{UGV Hardware}};
   
\end{tikzpicture}

%% file: state_machine.pgf
\begin{tikzpicture}[font=\sffamily,on grid,>={Stealth[inset=0pt,length=4pt,angle'=45]}]
\tikzset{every node/.append style={node distance=3.0cm}}
\tikzset{terminal_node/.append style={minimum size=1.5em,minimum height=3em,minimum width={width("Search Point")+0.2em},draw,align=center,rounded corners,scale=0.65}}
\tikzset{content_node/.append style={minimum size=1.5em,minimum height=3em,minimum width={width("Search Point")+0.2em},draw,align=center,scale=0.65,fill=blue!15!white}}
\tikzset{label_node/.append style={scale=0.5, near start}}
\tikzset{group_node/.append style={align=center,rounded corners,inner sep=1em,thick}}
\tikzset{decision_node/.append style={align=center,scale=0.5,shape aspect=1.5,minimum width=7.9em,minimum height=5.4em,diamond,draw,fill=yellow!25!white,font=\sffamily\normalsize,node distance=3.9cm}}

\definecolor{red}{rgb}     {0.5,0.0,0.0}
\definecolor{green}{rgb}   {0.0,0.5,0.0}
\definecolor{blue}{rgb}    {0.0,0.0,0.5}
\definecolor{grey}{rgb}    {0.5,0.5,0.5}

\draw[thick, rounded corners, grey!20!white,fill] (1, 3.2) -- (4.75, 3.2) -- (4.75, -0.8) -- (1, -0.8) -- cycle;
\draw[thick, rounded corners, grey!20!white,fill] (4.95, 3.2) -- (11.1, 3.2) -- (11.1, -0.8) -- (4.95, -0.8) -- cycle;

\node(start)[terminal_node,fill=red!15!white] at (0, 0) {Start};
\node(enter_building)[content_node, above of=start] {Enter Building};
\node(has_next_fire)[decision_node, right of=enter_building] {Has Next\\Location?};
\node(drive_to_fire)[content_node, below of=has_next_fire] {Drive to\\Location};
\node(perform_search_motion)[content_node, right of=drive_to_fire] {Perform Search\\Motion};
\node(has_thermal_detection)[decision_node, above of=perform_search_motion] {Has Thermal\\Detection?};

\node(aim_hose)[content_node, right of=has_thermal_detection] {Aim Hose};
\node(start_pump)[content_node, below of=aim_hose] {Start Pump};
\node(phase1)[decision_node, right of=start_pump] {Phase 1\\Complete?};
\node(spray_pattern)[content_node, right of=phase1] {Spray in\\Pattern};
\node(phase2)[decision_node, above of=spray_pattern] {Phase 2\\Complete?};
\node(stop_pump)[content_node, left of=phase2] {Stop Pump};
\node(stop)[terminal_node,fill=red!15!white,above of=has_next_fire] {Stop};

\draw[->, thick] (start) -- (enter_building);
\draw[->, thick] (enter_building) -- (has_next_fire);

\draw[->, thick] (has_next_fire) -- node[label_node,right] {Yes} (drive_to_fire);
\draw[->, thick] (has_next_fire) -- node[label_node,right] {No} (stop);
\draw[->, thick] (has_thermal_detection) -- node[label_node,above] {Yes} (aim_hose);
\draw[->, thick] (drive_to_fire) -- (perform_search_motion);
\draw[->, thick] (has_thermal_detection) -- node[label_node,above] {No} (has_next_fire);
\draw[->, thick] (has_thermal_detection) -- (aim_hose);
\draw[->, thick] (perform_search_motion) -- (has_thermal_detection);

\draw[->, thick] (aim_hose) -- (start_pump);
\draw[->, thick] (start_pump) -- (phase1);
\draw[->, thick] (phase1) -- node[label_node,above] {Yes} (spray_pattern);
\draw[->, thick] (phase1.north) -- node[label_node,right, midway] {No} ++(0, 0.4) -- ++(-1.1, 0) |- (phase1);
\draw[->, thick] (spray_pattern) -- (phase2);
\draw[->, thick] (phase2) -- node[label_node,above] {Yes} (stop_pump);
\draw[->, thick] (phase2.east) -- node[label_node,above, midway] {No} ++(0.4, 0) -- ++(0, -1.2) -| (phase2);
\draw[->, thick] (stop_pump) -- ++(0, 0.8) -- +(-5, 0) -- (has_next_fire);

\node[scale=0.65, anchor=north east] at (4.5, 3.2) {\textbf{Search Fire}};
\node[scale=0.65, anchor=north east] at (11, 3.2) {\textbf{Extinguish}};
\end{tikzpicture}

%% file: evaluation.tex
\begin{figure}[t]
  \centering
  \begin{tikzpicture}
    \node {\includegraphics[width=.9\linewidth]{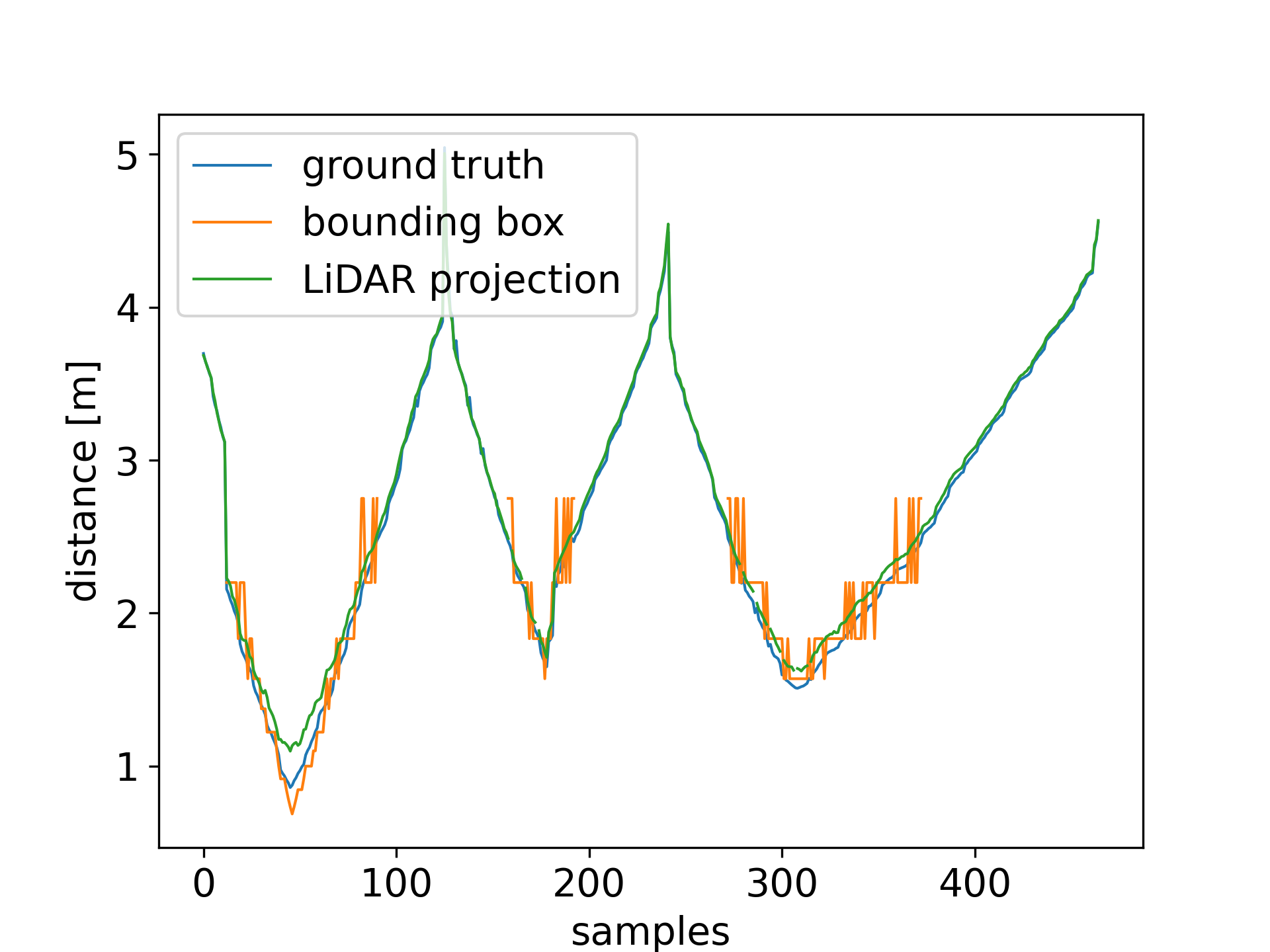}};
    \draw[red,line width=1pt] (-2.0cm, -1.0cm) circle [x radius=.7cm, y radius=1.2cm];
  \end{tikzpicture}
  \caption{Comparison of distance estimates from bounding box (\refsec{sec:ugv_perception}), LiDAR projection (\refsec{ssec:uav_thermal_detection}), and real distance.}
  \label{fig:distance_comparison}
\end{figure}
The two variants for estimating the position of the heat source described above differ in their application range. The LiDAR projection approach explained in \refsec{ssec:uav_thermal_detection} is required for distances above \SI{3}{\meter}, while the bounding box method described in \refsec{sec:ugv_perception} works best for distances smaller than \SI{2}{\meter}. \reffig{fig:distance_comparison} depicts a comparison of both methods to the ground truth distance obtained by fitting a plane to the test wall. The red encircled part indicates the benefits of the bounding box method for close distances. Estimates below \SI{1}{\meter} would be even more accurate, but the minimum distance of the LiDAR for acquiring ground truth data does not permit this measurement. The large jumps in the bounding box estimates above \SI{2}{\meter} originate from the discrete nature of the bounding boxes in the low-resolution thermal images. This figure does not compare the quality but shows that both variants perform well in their designated operating range.

\subsection{UAV}
\begin{figure}[t]
  \centering
  \begin{tikzpicture}
	[boxstyle/.style={orange!80!black,fill=orange!80!black,fill opacity=0.5,text opacity=1,text=black,draw,ultra thick,align=center}]
	\node[anchor=south east, inner sep = 0] (left_image) at (0,0) {\includegraphics[width=\linewidth]{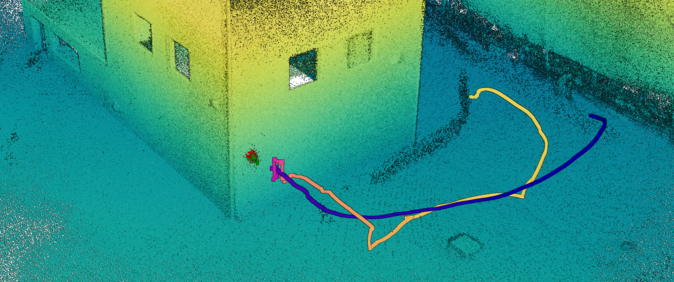}};
  \end{tikzpicture}
  \caption{Fire detection and UAV localization during manual fire extinguishing of the windy ground level facade fire during the second challenge day. The trajectory is color-coded by time (yellow to blue).}
  \label{fig:manual_fire}
\end{figure}

On the first scored challenge run, software issues and wrong predefined search poses prevented the UAV from detecting any fire. Similarly, we experienced incorrect height estimates on the second challenge day. We attribute this to the ultrasonic sensor measuring the building wall rather than the ground, thus estimating the height too low. After the Grand Challenge, we also noticed that during all trials, the LiDAR localization was disabled. As a result, the UAV flew too high to detect the fire. We then switched to manual mode and were able to fill the container of the windy ground-level facade fire with \SI{322}{\milli\liter} of water. The manually flown trajectory and localized detections of the fire are shown in~\reffig{fig:manual_fire}.

\begin{figure}[b]
  \centering
  \begin{tikzpicture}
	[boxstyle/.style={orange!80!black,fill=orange!80!black,fill opacity=0.5,text opacity=1,text=black,draw,ultra thick,align=center}]
	\node[anchor=south east, inner sep = 0] (left_image) at (0,0) {\includegraphics[width=\linewidth]{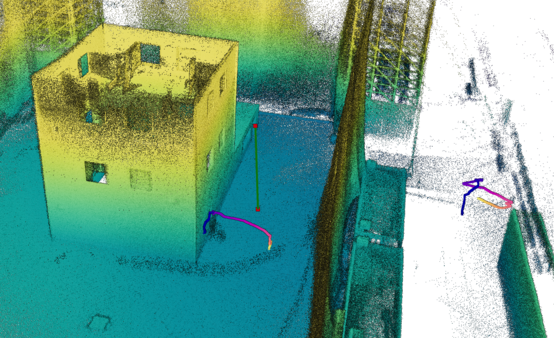}};
\begin{scope}[shift=(left_image.south west),x={(left_image.south east)},y={(left_image.north west)}]
        \node[boxstyle,text width=1.9cm] at (0.43,0.11) {Laser\\Localization};
        \node[boxstyle,text width=1.9cm] at (0.875,0.6) {GNSS-based\\Localization};
        \node[red,draw,line width=2pt,from={ 0.35,0.2 to 0.5,0.4}]{};
        \node[red,draw,line width=2pt,from={0.8,0.35 to 0.95,0.5}]{};
    \end{scope}
  \end{tikzpicture}
  \caption{Incorrect GNSS-based UAV localization during the Grand Challenge. The laser localization shows the trajectory before crashing into the building, color-coded by time (yellow to blue).}
  \label{fig:uav_crash_and_burn}
\end{figure}
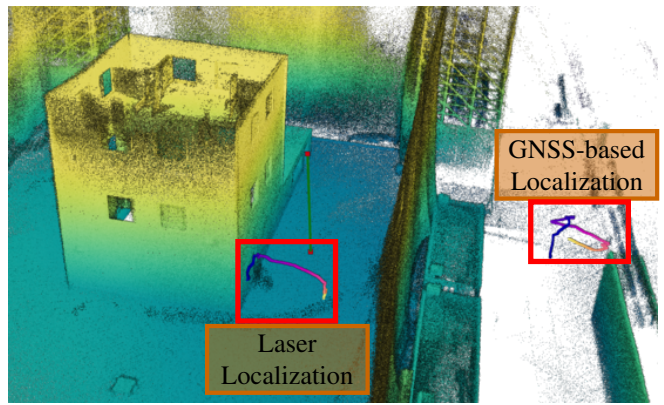

Shortly after the Grand Challenge started, a series of unfortunate circumstances led to the UAV crashing into the building, as shown in \reffig{fig:uav_crash_and_burn}. The laser localization was disabled due to human error, and position estimates relied solely on the GNSS-based ego-motion estimates from the UAV.
To compensate for position drift, we added a static offset to the GPS poses, which was calculated as the difference between the predefined start position and the GPS pose at the time of the challenge start. However, the GPS signal strongly drifted while computing the offset. Thus, a large incorrect offset was added to the GPS pose, resulting in an initial localization error of around \SI{20}{\meter}. Moreover, it continued to drift afterwards. The canyon-like starting position between the building and the operator booth, as well as interference with the UGV, might be responsible for the drift.

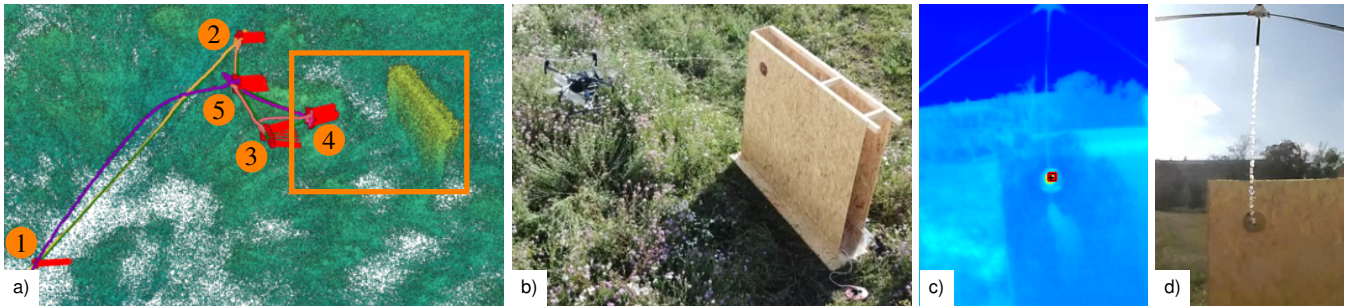
\begin{figure*}[t]
  \centering
  \begin{tikzpicture}
	[boxstyle/.style={orange!80!black,fill=orange!80!black,fill opacity=0.5,text opacity=1,text=black,draw,ultra thick,align=center}]
	\node[anchor=south east, inner sep = 0] (left_image) at (0,0) {\includegraphics[height=4cm]{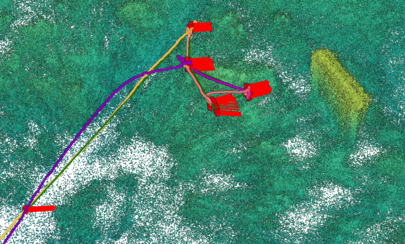}};
\begin{scope}[shift=(left_image.south west),x={(left_image.south east)},y={(left_image.north west)}]
        \node[fill=orange,circle,inner sep=0pt,minimum size = 12pt]at(0.035,0.21){1};
        \node[fill=orange,circle,inner sep=0pt,minimum size = 12pt]at(0.42,0.9){2};
        \node[fill=orange,circle,inner sep=0pt,minimum size = 12pt]at(0.495,0.495){3};
        \node[fill=orange,circle,inner sep=0pt,minimum size = 12pt]at(0.65,0.55){4};
        \node[fill=orange,circle,inner sep=0pt,minimum size = 12pt]at(0.43,0.65){5};
        \node[orange,draw,line width=2pt,from={ 0.57,0.37 to 0.92,0.83}]{};
    \end{scope}
    \node[anchor=south west,inner sep=0, right = .1cm of left_image] (right_image) {\includegraphics[height=4cm]{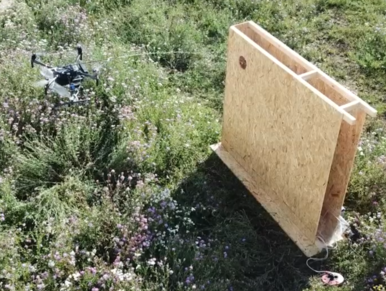}};
    \node[anchor=south west,inner sep=0, right = .1cm of right_image] (thermal_image) {\includegraphics[height=4cm]{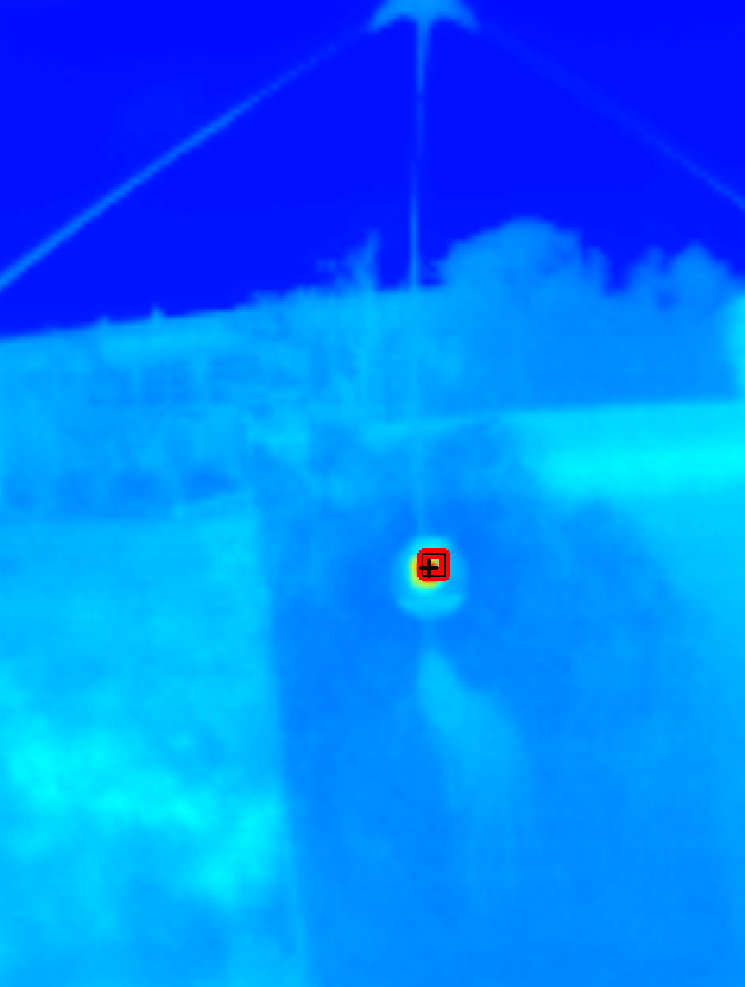}};
    \node[anchor=south west,inner sep=0, right = .1cm of thermal_image] (insta_image) {\includegraphics[height=4cm]{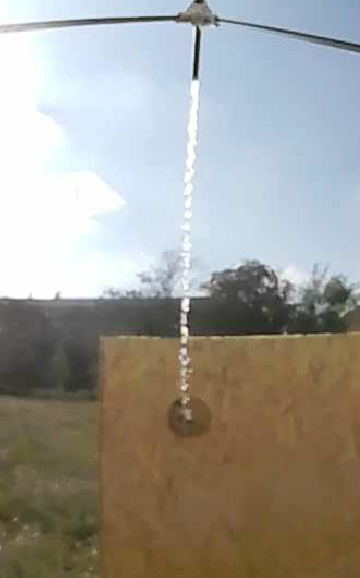}};
    \node[label,scale=1.0, anchor=south west, rectangle, fill=white, align=center, font=\scriptsize\sffamily] (n_1) at (left_image.south west) {a)};
    \node[label,scale=1.0, anchor=south west, rectangle, fill=white, align=center, font=\scriptsize\sffamily] (n_2) at (right_image.south west) {b)};
    \node[label,scale=1.0, anchor=south west, rectangle, fill=white, align=center, font=\scriptsize\sffamily] (n_3) at (thermal_image.south west) {c)};
    \node[label,scale=1.0, anchor=south west, rectangle, fill=white, align=center, font=\scriptsize\sffamily] (n_4) at (insta_image.south west) {d)};
  \end{tikzpicture}
  \caption{Autonomous fire extinguishing on mockup target with heating element. Red line segments indicate the UAV's heading at various numbered waypoints. When the heat source is detected [c)], the UAV approaches [orange rectangle, a)] and starts to extinguish [b-d)].}
  \label{fig:uav_poppelsdorf}
\end{figure*}

After repair, we performed further tests to showcase our UAV's ability with the mockup shown in \reffig{fig:uav_poppelsdorf}. After lift-off, the UAV flies towards multiple predefined search waypoints (\reffig{fig:uav_poppelsdorf} left, Poses~\num{1}-\num{3}). The heat source is first detected at Waypoint~\num{3}. Now the UAV begins to navigate relative to the detection and turns towards it before flying closer (Pose~\num{4}) and extinguishing the fire (\reffig{fig:uav_poppelsdorf} middle and right). After extinguishing, the UAV proceeds to Pose~\num{5}.

\subsection{UGV}

\reffig{fig:GrandChallengeUGV} shows an overview of the UGV's behavior during the scored runs. On the first challenge day, the robot could not compete due to software issues. During the second challenge run, the autonomy worked well to the point where the UGV tried to position the arm for scanning the first location for heat signatures. Because the manually configured approximate fire location was wrong (c.f.~\reffig{fig:GrandChallengeUGV} left), the arm collided with the target, which triggered an emergency stop. While manually maneuvering the robot out of the building during a reset, the not sufficiently tested motor controller malfunctioned, and the robot drove into a cupboard (light-pink trajectory). After resetting to the start zone, the faulty controller caused the robot to drive towards the arena boundary (magenta trajectory), even though localization and waypoints were correct. Near the end of the run, we switched to manual mode. With arm motions and a pump macro prepared in advance, the robot managed to spray \SI{470}{\milli\liter} water into the container just before the timeout.

After analyzing the collected data, we found localization inaccuracies while entering the building (see \reffig{fig:GrandChallengeUGV} left) due to the proximity of inside and outside points in the map. After creating separate maps and switching on driving inside, the localization performed accurately.

During the Grand Challenge run, the UGV successfully solved the task autonomously. There was an issue with a disconnected tf-tree, which led to the state machine stopping in front of the first fire. After a reset, we solved the issue, and the UGV successfully performed as intended: it stopped in front of the first location, scanned for the heat source, aimed and sprayed sufficient water to achieve the full score. \reffig{fig:GrandChallengeUGV} right shows driven trajectories during the first (purple) and second (magenta) attempt. The deviation after the first waypoint is due to an edge on the uneven ground; otherwise, the overlap of planned- and driven trajectories demonstrates our method's high accuracy.

A video showcasing the evaluation can be found on our website\footnote{\scriptsize{\url{https://www.ais.uni-bonn.de/videos/icuas_2021_mbzirc/}}}

%% file: lessons_learned.tex
Robotic competitions not only allow evaluation of the performing systems but also represent the opportunity to discover systematic problems and to learn from these.

As demonstrated by our problematic runs during the competition, it is highly challenging to develop such a complex system to a robustness level such that it immediately performs well in competition situations.  
Of course, having more testing time \textit{with a similar environment setup} as well as having proper scenario specifications early on with a limited number of sub-challenges would have helped this point immensely.

While we tested our UAV system with a mockup in our lab and on an open field, the competition environment was unexpectedly dissimilar and led to system failures.
For example, GPS visibility was severely limited near the mandatory start position, which led to initialization errors. An initialization-free or delayed initialization scheme, which we implemented later, or even a GPS drift detection would have improved robustness. Furthermore, relative navigation enables task fulfillment with unreliable pose information, and low-level obstacle avoidance is mandatory in such a scenario.
Furthermore, improved visualization of the UAV state and perception could have helped to detect problems early on.

The competition also placed an enormous strain on the involved personnel. For safe operation, one safety pilot was required per UAV, plus at least one team member supervising the internal state of the autonomous systems and being able to reconfigure the system during resets.
While reducing the number of required human operators per robot is an admirable research goal, this is not possible in the near future due to safety regulations. We thus feel that future competitions should keep the required human personnel in mind so that small teams can continue to participate.

UGV motor controller failures led to crashes in the competition and highlighted a lack of proper subsystem-level testing. The controllers were tested only with manual driving commands, where integrator wind-up is less likely to occur. The lack of testing with autonomous control led to the overloading of the responsible developers. In conclusion, our team size was too small for the numerous different tasks.

What proved to be useful in the context of competitions is to prepare low-effort backup strategies in advance, like arm motions for manual mode. Also, reducing high-level control to a bare minimum, as we did for the UGV with two fixed fire locations instead of a universal search-and-extinguish, makes the systems easier to test and reduces potential errors.

\begin{figure}[t]
  \centering
    \includegraphics[height=4.7cm]{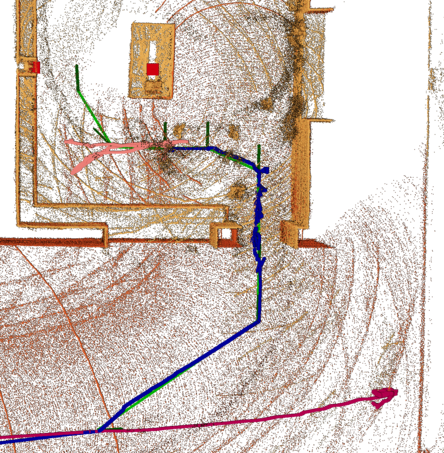}\hfill
    \includegraphics[height=4.7cm]{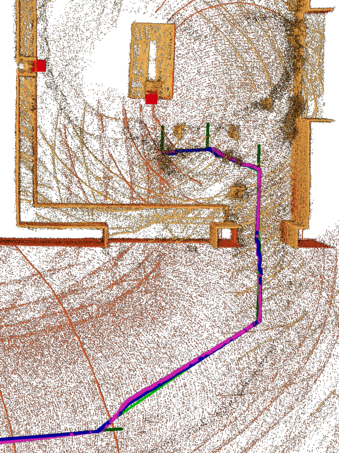}
  \caption{Navigation map and UGV trajectories on the second trial day (left) with active Fire~2 and the Grand Challenge run (right) with active Fire~1. The light green straight lines show the planned trajectory, where each branching dark green segment is a waypoint with the direction indicating the robot's heading. The other lines are the actual driven trajectory according to localization, colored differently after each reset. Two red rectangles predefine the suspected fire locations to scan for heat signatures.}
  \label{fig:GrandChallengeUGV}
\end{figure}

%% file: conclusion.tex
This paper presented an autonomous UAV-UGV team for fire extinguishing in Challenge 3 of the MBZIRC 2020. Our UGV Bob successfully delivered the maximum amount of water into the fire in the Grand Challenge. Even though our UAV did not work during the competition, after analyzing and fixing the issues, our experiments show that the system functions reliably.

The shared LiDAR localization works accurately on both platforms with different LiDAR sensors. This was especially important to maneuver the oversized UGV chassis, initially developed for Challenge 2, through the door and within the restricted space inside the building. The UGV high-level control was tailored to the scenario with two fixed fire locations but can be easily extended by driving an exploratory path while scanning the environment for heat sources with continuous arm motions. The methodology proved to be sound, and failures during the competition originated from insufficient testing and last-minute changes.

We detected the fire targets from complementary modalities and fused geometric hole features from LiDAR with heat detections from thermal cameras. Our filter allowed us to approach the fire, while both robots utilize relative navigation to reliably and repeatedly spray water into the fire.

Aside from analyzed and corrected issues during the competition, we showed that both robots can perform their designated tasks autonomously and that the perception, motion planning, and fire extinguishing works reliably. The knowledge gained and components developed will also prove useful in future competitions.